\documentclass[runningheads]{llncs}
\usepackage{enumitem}

\usepackage{eccv}
\usepackage{eccvabbrv}
\usepackage{graphicx}
\usepackage{booktabs}
\usepackage{array}
\usepackage{caption}
\usepackage{subcaption}
\graphicspath{{images/}}

\usepackage[accsupp]{axessibility}
\usepackage{hyperref}
\usepackage{orcidlink}
\begin{document}
\title{Why 3D Reconstruction Breaks Down on Under-Ice AUV Video, and How to Predict It}
\titlerunning{Predicting Under-Ice Reconstruction Failure}

\author{Ana-Ștefania Kogălniceanu\inst{1} \and
Klára Orbán\inst{1} \and
Ashwin Nedungadi\inst{2} \and
Stefan Lüdtke\inst{2} \and
Kuderna-Iulian Bența\inst{1}}

\authorrunning{A.-Ș. Kogălniceanu et al.}

\institute{Babeș-Bolyai University, Cluj-Napoca, Romania\\
\email{\{ana.kogalniceanu,klara.orban\}@stud.ubbcluj.ro},
\email{kuderna.benta@ubbcluj.ro}
\and
University of Rostock, Germany\\
\email{\{ashwin.nedungadi,stefan.luedtke\}@uni-rostock.de}}
\maketitle

\begin{abstract}
Reconstruction of 3D structures using under-ice AUV video footage has been inaccurate, although it is not certain whether this is due to shortcomings in the methods or an inherent deficiency of the capture system. We test four different models for reconstruction on the same Antarctic under-ice AUV video (Icefin, Kamb Ice Stream): classical feature matching SfM, differentiable photometric splatting, volumetric depth fusion, and a pose-free feed-forward foundation model. All four fail, each in a distinct way: near-total registration yielding almost no structure, geometrically broken output despite healthy photometric metrics, a persistent coverage artifact, and unstable cross-chunk scale. We show that all four are consistent with the same capture regime, a small camera-baseline-to-scene-depth ratio. Following our analysis, we propose a detection system that flags this type of failure in minutes, rather than hours of wasted compute.

  \keywords{under-ice 3D reconstruction \and autonomous underwater vehicles \and narrow-baseline photogrammetry \and 3D Gaussian Splatting \and texture-usability estimation}
\end{abstract}

\section{Introduction}
\label{sec:introduction}

3D reconstruction from under-ice AUV video is persistently unreliable: turbidity and backscatter flatten contrast, while the ice underside is often smooth and weakly textured \cite{akkaynak2018revised,li2019seaicesfm}. Although mitigations exist, it remains unclear whether the residual failures reflect the shortcomings of any one method or a deeper limitation of the capture geometry itself. Slow-moving AUVs sampled at low frame rates produce near-converging camera trajectories with limited translation between viewpoints, unlike the wide-baseline, well-distributed views on which most reconstruction pipelines are designed and validated. In practice, researchers face large collections of such footage where reconstruction success is difficult to anticipate, and substantial compute is routinely spent on batches that converge to geometrically fragmented output, with the failure only apparent after the fact.

Prior work in this space has largely evaluated individual pipelines in isolation
\cite{colmap,3dgs,vggt}, without asking whether failures across structurally
different methods share a common, measurable cause. This distinction matters: if
failure is method-specific, the fix is method-specific tuning; if failure is
geometric, the fix is a single early warning system that generalizes across
pipelines.

This work tests four structurally unrelated reconstruction approaches: classical feature-matching SfM, differentiable photometric splatting, volumetric depth fusion, and a pose-free feed-forward foundation model -- on the same under-ice AUV footage (Icefin, Kamb Ice Stream, USAP-DC 601975 \cite{washam2025icefin}), and shows that all four fail for reasons traceable to the same capture regime. The main contributions of this work are:
\begin{itemize}[label=\textbullet]
    \item identification of a single quantified geometric predictor: a narrow camera baseline-to-scene-depth ratio that explains the distinct failure modes of four structurally different reconstruction methods on the same footage,
    \item a lightweight baseline-to-depth check, derived directly from foundation-model camera poses, that flags likely failure in minutes rather than hours,
    \item a negative control against the competing hypothesis that reconstructability is predictable from image content, rather than capture geometry: a texture-usability score, tested against altimeter ground truth, did not hold up.
\end{itemize}

\section{Related work}
\label{sec:related_work}

Classical Structure-from-Motion methods such as COLMAP \cite{colmap} recover camera poses and sparse structure through incremental feature matching and bundle adjustment, with dense surfaces typically obtained through volumetric depth fusion into a signed-distance field \cite{curless1996tsdf}. While these pipelines are well validated for wide-baseline, well-distributed captures, their behavior under narrow-baseline, near-converging viewpoints remains comparatively unexplored. 3D Gaussian Splatting (3DGS) \cite{3dgs} replaces explicit correspondence search with differentiable photometric optimization, achieving high-quality novel-view synthesis without directly guaranteeing geometric consistency, a limitation that underwater adaptations have had to address \cite{uw3dgs}. Feed-forward foundation models such as DUSt3R \cite{dust3r} and its more scalable successor VGGT \cite{vggt} instead predict camera poses and dense point maps directly from images, avoiding iterative optimization but introducing a different failure mode: maintaining consistent scale and pose across inference chunks.

AUV-based seafloor mapping has similarly reported reconstruction gaps and weakly reconstructed regions caused by platform motion and uneven illumination \cite{she2024seafloor}. Icefin-based surveys demonstrate the scientific value of under-ice observation \cite{icefin_kamb}, but the reliability of 3D reconstruction from such footage remains largely unaddressed. A smaller body of work has explored predicting reconstruction quality before running a full pipeline, including models linking UAV image characteristics such as blur, exposure, and noise to point-cloud accuracy \cite{uav_error_pred}, no-reference image-quality measures based on photogrammetric descriptors \cite{keypoint_iqa}, and preprocessing strategies for blurred or texture-poor imagery \cite{preprocessing_arch}. However, these approaches primarily assess 2D image quality rather than 3D capture geometry.

\section{Dataset and Capture Setup}
\label{sec:dataset}

We use video footage collected by the Icefin autonomous underwater vehicle (AUV) during dive ICE03\_037 in the Kamb Ice Stream, Antarctica , available through the U.S. Antarctic Program Data Center (USAP-DC 601975) \cite{washam2025icefin}. The dataset consists of downward- and forward-facing HD and SD video recorded as the vehicle moved slowly through the subglacial ice channel. For reconstruction experiments, frames were sampled at 1--4 fps from the \texttt{box1} and \texttt{box6} video clips, providing sequences with varying amounts of usable ice-surface texture. The footage presents a challenging reconstruction setting because the vehicle moves slowly and predominantly toward the observed surface, producing short camera baselines relative to scene depth and near-converging viewpoints.

\section{Convergent Failure Across Four Methods}
\label{sec:experimental_methods}

To test whether the reconstruction limitations were algorithm-specific, we evaluated
the same under-ice footage using four structurally different approaches. Despite
substantially different failure modes, all four are consistent with one capture regime: insufficient
camera baseline relative to scene depth.

\subsection{Classical SfM: Registration Without Reconstruction}

We first evaluated COLMAP \cite{colmap} on a 1~fps sequence of 62 frames from the \texttt{box6} clip; exhaustive matching registered only 2 of 62 images.
Increasing the sampling rate to 4~fps and using COLMAP's sequential matcher with
loop closure substantially improved registration, with 247 of 248 frames
successfully registered. Registration rate, however, measures completeness
rather than reconstruction quality: the model contains only 817 points across
248 images ($3.3$/image, versus $40.4$/image for \texttt{box1} processed
identically), with a median track length of $4$, so most features persist across
only a handful of near-identical consecutive frames rather than being observed
from distinct viewpoints. Mean reprojection error is $0.76$~px: the bundle
adjustment is well satisfied, but by a sparse set of correspondences that does
not constrain scene geometry. Denser sampling recovers registration without
recovering structure.

\subsection{Photometric Splatting: Healthy Metrics, Broken Geometry}

We next trained 3D Gaussian Splatting (3DGS) \cite{3dgs} using the COLMAP reconstruction, so 3DGS inherits COLMAP's camera poses rather than constituting an independent pose estimate. The resulting model was applied on \texttt{box6} and achieved a test PSNR of 24.75 after 30,000 iterations, suggesting that the model reproduced the training-view appearance effectively. However, this metric concealed severe geometric degradation. When the reconstruction was rendered from a free viewpoint outside the original camera trajectory, the scene contained elongated, needle-like Gaussians extending away from the observed surface. The result demonstrates that photometric agreement from a narrow range of viewpoints does not necessarily imply geometrically valid reconstruction: with insufficient baseline, degenerate geometry can reproduce the observed images without being constrained by meaningful multi-view triangulation.

\subsection{Depth Fusion: Bounded by Real Overlap}

To test whether replacing photometric optimization with explicit depth fusion avoided the problem, we estimated per-frame depth with Depth Anything~3 (DA3) \cite{da3} in monocular mode, back-projected the estimates into the reconstructed camera coordinate system, and fused them volumetrically using a TSDF. An initial configuration produced almost no consolidation, since the truncation band was too narrow relative to the disagreement between depth maps. Widening the voxel size and truncation distance reduced the surface from 15.0 million to 667,000 vertices, a substantial consolidation, yet the characteristic fan-like silhouette remained. Monocular depth carries per-frame scale and shift ambiguity, which contributes to this disagreement independently of baseline; DA3's pose-conditioned and multi-view modes would separate the two contributions, and we leave that comparison to future work.

\begin{figure}[tb]
  \centering
  \begin{subfigure}{0.32\linewidth}
    \centering
    \includegraphics[height=3cm,keepaspectratio]{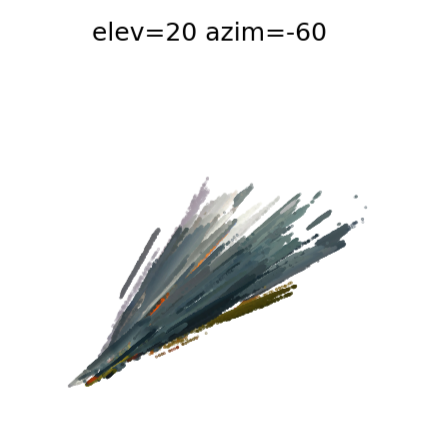}
  \end{subfigure}
  \hfill
  \begin{subfigure}{0.32\linewidth}
    \centering
    \includegraphics[height=3cm,keepaspectratio]{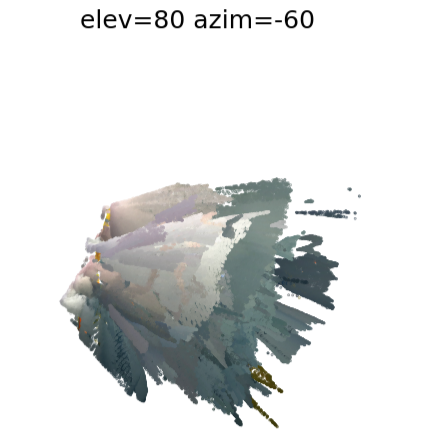}
  \end{subfigure}
  \hfill
  \begin{subfigure}{0.32\linewidth}
    \centering
    \includegraphics[height=3cm,keepaspectratio]{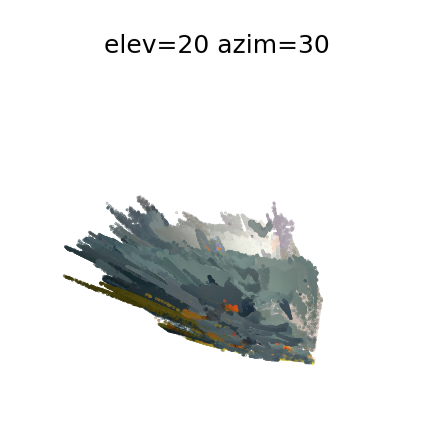}
  \end{subfigure}
  \caption{TSDF fusion of \texttt{box1} after increasing voxel size and
  truncation distance, reducing the surface from 15.0M to 667K vertices
  ($22\times$ consolidation). The fan-like silhouette persists, indicating a
  limitation of viewpoint overlap rather than fusion parameters.}
  \label{fig:tsdf}
\end{figure}

\subsection{Foundation Models: Local Success, Global Failure}

Finally, we evaluated VGGT \cite{vggt}, which predicts camera poses and dense point maps directly from images without iterative feature matching or bundle adjustment. On the \texttt{box6} frame sequence for which COLMAP initially registered only two images, VGGT produced valid local reconstructions for all 62 frames. Individual inference chunks showed coherent geometry, demonstrating that the model can extract useful local structure even when classical feature matching fails. However, independently processed chunks could not be reliably combined into a common reconstruction. Similarity alignment of adjacent chunks produced relative scale factors ranging from 0.43 to 1.73, rather than the approximately unit scale expected for consistent reconstructions. Increasing the overlap between chunks did not resolve this instability, indicating that the problem is not simply insufficient overlap redundancy but the lack of sufficient translational baseline to constrain relative scale.

\begin{figure}[tb]
  \centering
  \begin{subfigure}{0.24\linewidth}
    \includegraphics[width=\linewidth]{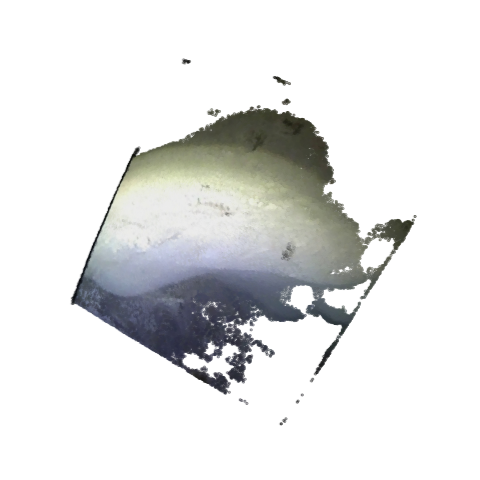}
    \caption{Single chunk}
  \end{subfigure}\hfill
  \begin{subfigure}{0.24\linewidth}
    \includegraphics[width=\linewidth]{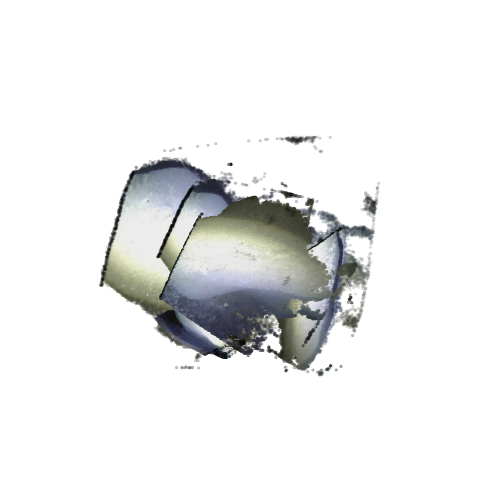}
    \caption{Four chunks merged}
  \end{subfigure}\hfill
  \begin{subfigure}{0.24\linewidth}
    \includegraphics[width=\linewidth]{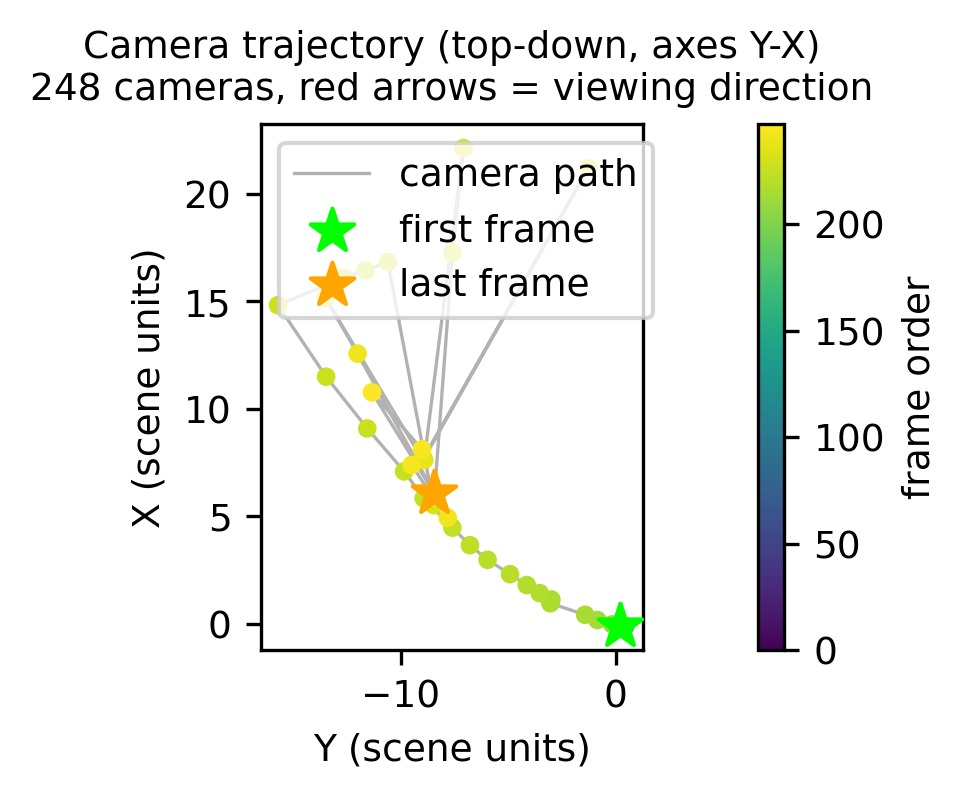}
    \caption{Camera centers}
  \end{subfigure}\hfill
  \begin{subfigure}{0.24\linewidth}
    \includegraphics[width=\linewidth]{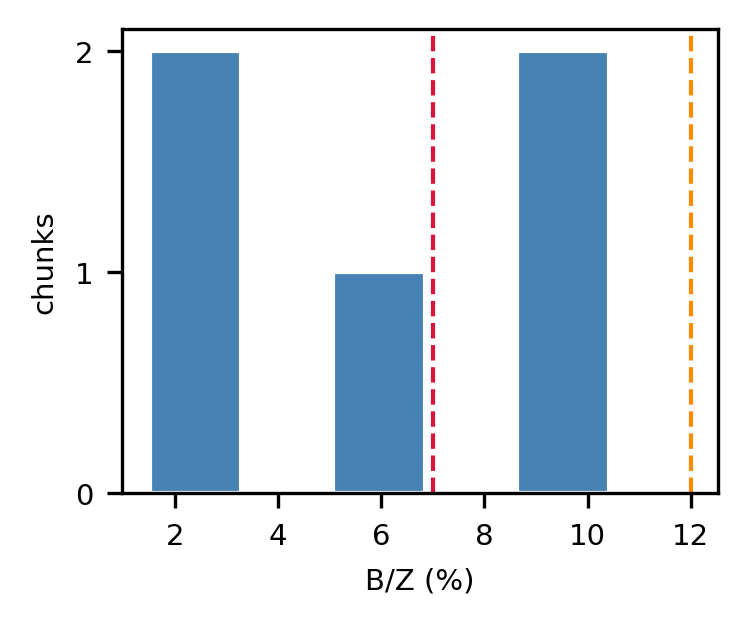}
    \caption{Per-chunk $B/Z$}
  \end{subfigure}
  \caption{VGGT on the 62-frame \texttt{box6} sequence. Chunks are locally
  coherent (a) but merge inconsistently, with scale factors
  $0.43$--$1.73\times$ (b). The recovered trajectory (c) advances predominantly
  toward the surface. Per-chunk baseline-to-depth ratios (d), with the 7\% and
  12\% tier thresholds marked (dashed).}
  \label{fig:vggt}
\end{figure}

\subsection{Camera-Pose Quality Control}
\label{sec:pose_qc}

The different failure modes become consistent when considered in terms of the
recovered camera trajectory. Across the evaluated sequences, the maximum
camera-center translation within VGGT inference chunks was only approximately
1--12\% of the scene depth, with most measurements below 10\%
(Figure~\ref{fig:vggt}c). This narrow baseline-to-depth ratio is therefore a useful diagnostic description of the capture regime associated with failure in our experiments: COLMAP recovers
few well-triangulated points, 3DGS can satisfy photometric observations using
degenerate geometry, depth fusion cannot force independently estimated surfaces
to agree, and VGGT cannot reliably determine a common scale across local
reconstructions.

One confound should be noted. Icefin's lighting is rigidly coupled to the
camera, so irradiance at a fixed surface point varies with range, violating the
static-illumination assumption all four pipelines share. This covaries with the
geometry we measure -- forward-dominant motion drives both -- and our data
cannot separate them, though VGGT's scale drift is not readily explained by
illumination, since relative scale is fixed by triangulation. We therefore treat
baseline-to-depth as a proxy for the capture regime rather than an isolated
cause.

\section{A Pre-Flight Baseline-Depth Check}
\label{sec:preflight}

Section~\ref{sec:experimental_methods} identified camera baseline-to-scene-depth ratio as a common failure-associated characteristic across the evaluated methods, computable from VGGT camera poses alone without dense reconstruction or ground truth. We therefore use the ratio as a practical diagnostic for the failure regime observed in our experiments, while leaving broader thresholds for future validation. Reconstruction failed throughout the full 1--12\% range
observed on our footage, so we anchor only a red tier (below 7\%, where failures
cluster densely) to evidence; the 7--12\% yellow and $\geq$12\% green tiers are
untested extrapolations, not validated thresholds.Computing the ratio takes $\sim$4 minutes (VGGT runtime), versus hours to discover the failure through COLMAP, 3DGS, or TSDF fusion. \texttt{box6} (measured 1.5--10.4\%
per chunk) falls in the red tier, consistent with its observed failure.

\section{Discussion and Limitations}
\label{sec:discussion}

We also tested whether reconstructability could be predicted from image content
rather than capture geometry. A texture-usability score showed negligible
agreement with the internal validators it was meant to complement (Spearman
$r=0.094$, $n=155$, $p=0.243$) and correlated with self-consistency in the wrong
direction ($r=-0.54$ to $-0.64$ across three formulations, $n=155$,
$p<0.001$).\footnote{These 155 frames come from a single \texttt{box6} clip,
processed as six DA3 inference batches; self-consistency is computed jointly
within a batch while confidence is computed per image, so the two are not
strictly comparable draws.} Against altimeter ground truth\footnote{391
altimeter-matched frames from five windows within a single dive (\texttt{box1}
and \texttt{box6}); not independent samples, hence the random effect below.},
the pooled correlation with DA3 depth error was weak ($r=-0.095$, $n=391$,
$p=0.061$); re-fitting as a linear mixed-effects model with window as a random
effect \cite{bates2015lme4} left the effect indistinguishable from noise
($p=0.838$). On this footage the binding constraint is capture geometry, not
image content.

Several limitations remain: evidence from a single dive, so the tier thresholds
are anchored to one platform and ice regime; a pinhole model throughout, though
Icefin's flat port makes the imaging refractive; VGGT's relative pose accuracy
unvalidated under near-axial motion; and altimeter data used only to test the
texture score, not to evaluate reconstruction accuracy metrically. We also lack
a positive control, footage that passes the check and reconstructs successfully.

\section{Conclusion}
\label{sec:conclusion}

Four structurally unrelated reconstruction methods -- classical SfM, photometric
splatting, volumetric depth fusion, and a pose-free feed-forward model -- fail on
the same under-ice AUV footage in four distinct ways, with all four experiments exhibiting a camera baseline of only 1--12\% of scene depth. Because this is computable from feed-forward camera poses alone, it supports a pre-run check that flags this footage as unreconstructable in minutes rather than hours, though only the failure tier is anchored to evidence here. Our  attempt to predict reconstructability from image content instead did not survive validation against altimeter ground truth -- on this footage, the
binding constraint is where the camera went, not what it saw. Extending the check across more dives, and deploying it onboard during a survey
so that flagged geometry can be corrected while the vehicle is still in the
water, remains future work.

\bibliographystyle{splncs04}
\bibliography{main}
\end{document}